\def\year{2020}\relax
\documentclass[10pt, journal, compsoc]{IEEEtran} % DO NOT CHANGE THIS
\usepackage{times}  % DO NOT CHANGE THIS
\usepackage{helvet} % DO NOT CHANGE THIS
\usepackage{courier}  % DO NOT CHANGE THIS
\usepackage[hyphens]{url}  % DO NOT CHANGE THIS
\usepackage{graphicx} % DO NOT CHANGE THIS
\usepackage{amsmath}
\usepackage{nicefrac}
\usepackage[super]{nth}
\usepackage{bm}
\usepackage{upgreek}
\usepackage{algorithm}
\usepackage{caption}
\usepackage{algpseudocode}
\usepackage{amssymb}
\usepackage{xcolor}
\usepackage{enumitem}
\algnewcommand\algorithmicforeach{\textbf{for each}}
\algdef{S}[FOR]{ForEach}[1]{\algorithmicforeach\ #1\ \algorithmicdo}
\algrenewcommand\algorithmicindent{1.0em}

\DeclareMathOperator*{\E}{\mathbb{E}}

\usepackage{booktabs}
\usepackage{multirow}
\usepackage{color, colortbl}
\usepackage{tabularx}
\usepackage{makecell}

\definecolor{Gray}{gray}{0.8}

\definecolor{tableBlue}{RGB}{189, 215, 238}
\newcommand{\CB}{\cellcolor{tableBlue}}

\definecolor{tableGreen}{RGB}{198, 224, 180}
\newcommand{\CG}{\cellcolor{tableGreen}}

\definecolor{tableOrange}{RGB}{255, 230, 153}
\newcommand{\CO}{\cellcolor{tableOrange}}

\newcommand{\MCR}[3]{\multicolumn{#1}{c}{\multirow{#2}{*}{#3}}}
\newcolumntype{P}{>{\centering\arraybackslash}p{1.1cm}}

\usepackage{graphicx}  				% DO NOT CHANGE THIS
\title{Multi-Task Federated Learning for Personalised \\ Deep Neural Networks in Edge Computing}

\begin{document}

\author{Jed Mills, Jia Hu, Geyong Min
	\IEEEcompsocitemizethanks{
	
		\IEEEcompsocthanksitem{ J. Mills, J. Hu and G. Min are with the 
		Department of Computer Science, University of Exeter, EX4 4QF, United Kingdom. 
		E-mail: \{jm729, j.hu, g.min\}@exeter.ac.uk. Corresponding authors: Jia Hu, Geyong Min. }% <-this % stops a space
		
		\IEEEcompsocthanksitem{This paper has been accepted by IEEE Transactions on Parallel and Distributed Systems.}
		\IEEEcompsocthanksitem{Source code for reproducing experiments on GitHub: https://github.com/JedMills/MTFL-For-Personalised-DNNs.}
	}
}

\IEEEtitleabstractindextext{%
\begin{abstract}
Federated Learning (FL) is an emerging approach for collaboratively training Deep Neural Networks (DNNs) on mobile devices, without private user data leaving the devices. Previous works have shown that non-Independent and Identically Distributed (non-IID) user data harms the convergence speed of the FL algorithms. Furthermore, most existing work on FL measures global-model accuracy, but in many cases, such as user content-recommendation, improving individual User model Accuracy (UA) is the real objective. To address these issues, we propose a Multi-Task FL (MTFL) algorithm that introduces non-federated Batch-Normalization (BN) layers into the federated DNN. MTFL benefits UA and convergence speed by allowing users to train models personalised to their own data. MTFL is compatible with popular iterative FL optimisation algorithms such as Federated Averaging (FedAvg), and we show empirically that a distributed form of Adam optimisation (FedAvg-Adam) benefits convergence speed even further when used as the optimisation strategy within MTFL. Experiments using MNIST and CIFAR10 demonstrate that MTFL is able to significantly reduce the number of rounds required to reach a target UA, by up to $5\times$ when using existing FL optimisation strategies, and with a further $3\times$ improvement when using FedAvg-Adam. We compare MTFL to competing personalised FL algorithms, showing that it is able to achieve the best UA for MNIST and CIFAR10 in all considered scenarios. Finally, we evaluate MTFL with FedAvg-Adam on an edge-computing testbed, showing that its convergence and UA benefits outweigh its overhead.
\end{abstract}

}

\maketitle 

\section{Introduction}
\IEEEPARstart{M}{ulti}-access Edge Computing (MEC) \cite{MECSurvey} moves Cloud services to the network edge, enabling low-latency and real-time processing of applications via content caching and computation offloading \cite{CompOffMEC} \cite{LearnAidCompOff}. Coupled with the rapidly increasing quantity of data collected by smartphones, Internet-of-Things (IoT) devices, and social networks (SNs), MEC presents an opportunity to store and process huge quantities of data at the edge, close to their source.

Deep Neural Networks (DNNs) for Machine Learning (ML) are becoming increasingly popular for their huge range of potential applications, ease of deployment, and state-of-the-art performance. Training DNNs in supervised learning, however, can be computationally expensive and require an enormous amount of training data, especially with the trend of increasing DNN size. The use of DNNs in MEC has typically involved collecting data from mobile phones/IoT devices/SNs, performing training in the cloud, and then deploying the model at the edge. Concerns about data privacy, however, mean that users are increasingly unwilling to upload their potentially sensitive data, raising the question about how these models will be trained. 

Federated Learning (FL) \cite{FedLearnChallenges} opens new horizons for ML at the edge. In FL, participating clients collaboratively train an ML model (typically DNNs), without revealing their private data. McMahan \textit{et al.} \cite{FedAvgPaper} published an initial investigation into FL with the \textit{Federated Averaging} (FedAvg) algorithm. FedAvg works by initialising a model at a coordinating server before  distributing this model to clients. These clients perform a round of training on their local datasets and push their new models to the server. The server averages these models together before sending the new aggregated model to the clients for the next round of training. \textcolor{black}{We refer to the people/institutions/etc. that own data for FL as `users', and to the devices that actually participate in FL as `clients'.}

FL is a very promising approach for distributed ML in situations where data cannot be uploaded for protecting clients' privacy. Therefore, FL is well suited for real-world scenarios such as analysing sensitive healthcare data \cite{BrainTorrent} \cite{FLPatientClustering}, next-word prediction on mobile keyboards \cite{GBoard}, and content-recommendation \cite{FLReccomendation}. However, FL presents multiple unique challenges:

\begin{itemize}[topsep=1pt]
\item{Clients usually do not have Independent and Identically Distributed (IID) training data. Each client has data generated by itself, and can have noisy data or only a subset of all features/labels. These factors can all substantially hinder training of the FL model.}

\item{\textcolor{black}{FL research typically uses the performance metric of global-model accuracy on a centralised test-set. However, in many cases, individual model accuracy on clients is the real objective - motivating `personalised FL' that creates unique models for FL clients to improve local performance. However, the best way of incorporating personalisation into FL remains an under-researched topic.}}

\item{Due to the non-IID nature of client datasets, the performance of the global FL model may be higher on some clients than others. This could even lead some clients to receive a worse model than the one they could have trained independently.}
\end{itemize}

\textcolor{black}{This paper addresses the above challenges by proposing a Multi-Task FL algorithm (MTFL), that allows clients to train personalised DNNs that both improve local model accuracy, and help to further enhance client privacy. MTFL has lower storage cost of personalisation, and lower computing cost compared with other personalised FL algorithms (not requiring extra steps of SGD on clients during the training loop or at personalisation time) \cite{Moreau} \cite{ImproveFLMAML} \cite{PerFLMAML} \cite{FedMultiTaskLearn}. }

As client datasets in FL are usually non-IID, clients can be viewed as attempting to optimise their models during local training for disparate tasks. Our MTFL approach takes the Batch-Normalisation (BN) layers that are commonly incorporated into DNN architectures, and keeps them private to each client. Mudrarkarta \textit{et al.} \cite{KFor1} previously showed that private BN layers improved Multi-Task Learning (MTL) performance for joint training on ImageNet/Places-365 in the centralised setting.

Using private BN layers has the dual benefit of personalising each model to the clients' local data as well as helping to preserve data privacy: as some parameters of client models are not uploaded to the server, less information about a client's data distribution can be gleaned from the uploaded model. Our MTFL approach using BN layers also has a storage-cost benefit compared to other personalised FL algorithms: BN layers typically contain a tiny fraction of the total parameters of a DNN, and only these BN parameters need to be stored between FL rounds, compared to entire personalised DNN models of competing algorithms \cite{Moreau} \cite{FedMultiTaskLearn} \cite{PerFLMAML}. 

MTFL adds personalisation on top of the typical iterative FL framework. FedAvg and other popular algorithms are instances of this iterative optimisation framework \cite{FedAvgPaper} \cite{SNIPSPaper} \cite{AdaptiveFedOpt}. Most of these FL algorithms use vanilla Stochastic Gradient Descent (SGD) on clients, however, momentum-based optimisation strategies such as Adam \cite{Adam} have the potential to improve convergence speed of FL training. We show that a distributed optimisation technique using Adam (FedAvg-Adam) shows substantial speedup in terms of communication rounds compared to FedAvg, and works very well within the MTFL algorithm.

Our work makes the following contributions: 

\begin{itemize}[topsep=0.25em]
\setlength\itemsep{0.25em}
\item{We propose an MTFL algorithm that adds Multi-Task learning on top of general iterative-FL algorithms, allowing users to learn DNN models that are personalised for their own data. MTFL uses private Batch Normalisation (BN) layers to achieve this personalisation, which provides an added privacy benefit.}

\item{We propose a new metric for measuring the performance of FL algorithms: User model Accuracy (UA). UA better reflects a common objective of FL (increasing test accuracy on clients), as opposed to the standard global-model accuracy.}

\item{We analyse the impact that private BN layers have on the activations of MTFL models during inference, providing insights into the source of their impact. We also analyse the training and testing performance of MTFL when keeping either the trained parameters or statistics of BN layers private, demonstrating that MTFL provides a better balance between convergence and regularisation compared to FL or independent training.}

\item{We conduct extensive simulations on the MNIST and CIFAR10 datasets. The results show that MTFL with FedAvg is able to reach a target UA in up to $5\times$ less rounds than when using only FL, with FedAvg-Adam providing a further $3\times$ improvement. Other experiments show that MTFL is able to significantly improve average UA compared to other state-of-the-art personalised FL algorithms.}

\item{We perform experiments using an MEC-like testbed consisting of Raspberry Pi clients and a FL server. The results show that MTFL with FedAvg-Adam's overheads are outweighed by its substantial UA and convergence speed benefits.}
\end{itemize}

The rest of this paper is organised as follows: Section 2 describes related work; Section 3 details the proposed MTFL algorithm, the effect that keeping private BN layers within MTFL has on training and inference, and the FedAvg-Adam optimisation strategy; Section 4 presents and discusses experiments using both simulations and an MEC-like testbed; and Section 5 concludes the paper.

\begin{figure*}[h]
\centering
\includegraphics[width=0.7\textwidth]{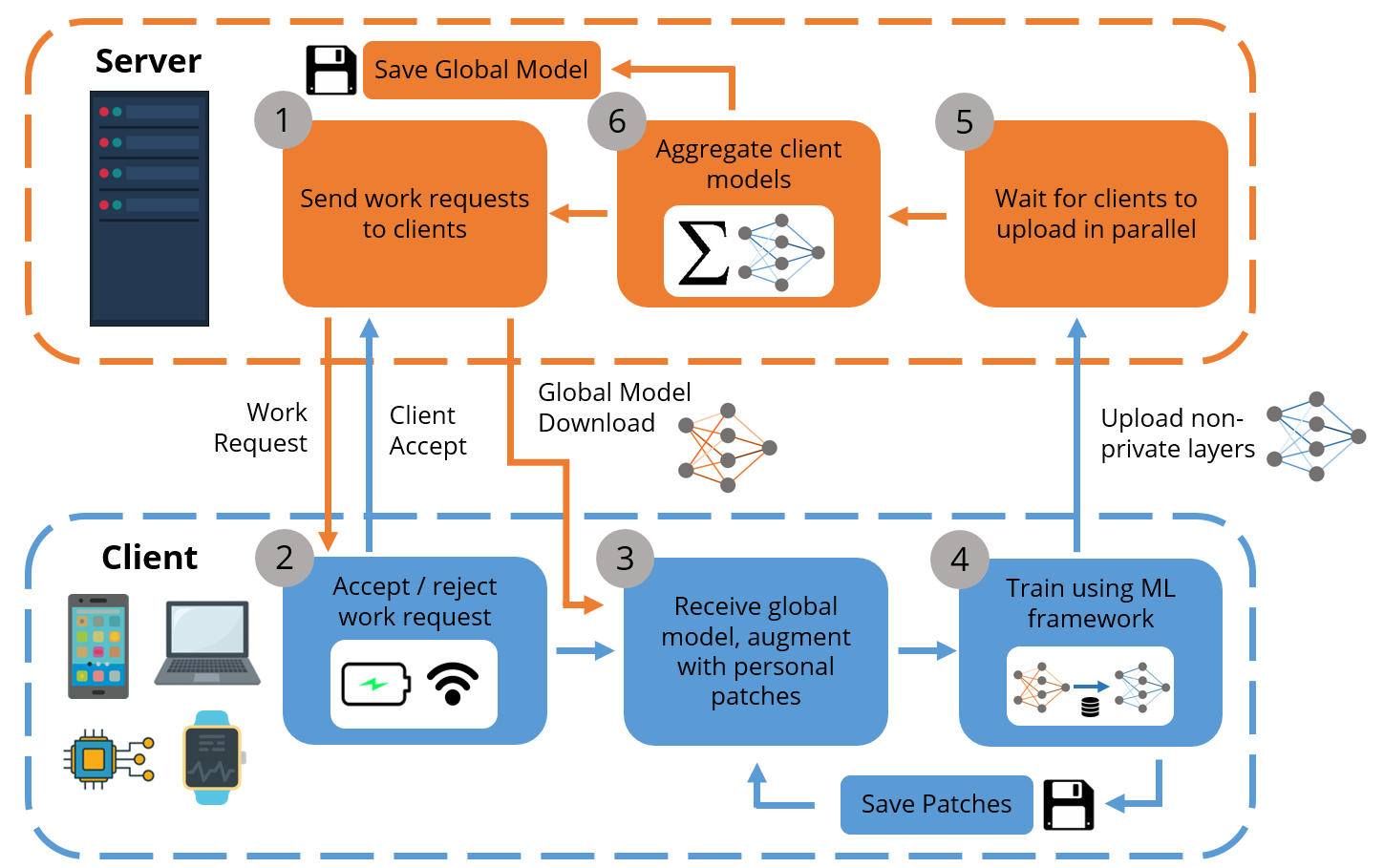}
\caption{Operation of the MTFL algorithm in Edge Computing. Training is performed in rounds until a termination condition is met. \textbf{Step 1}: the server selects a subset of clients from its database to participate in the round, and sends a work request to them. \textbf{Step 2}: clients reply with an accept message depending on physical state and local preferences. \textcolor{black}{\textbf{Step 3}: clients download the global model (and any optimisation parameters) from the server, and update their copy of the global model with private patches (in this work, we use BN layers as patches).} \textbf{Step 4}: clients perform local training, before saving their personal patches for the next round. \textbf{Step 5}: the server waits for $C$ fraction of clients to upload their non-private model and optimiser values, or until a time limit. \textbf{Step 6}: the server averages all models, saves the aggregate, and starts a new round.}
\label{fig1}
\end{figure*}

\section{Related Work}
As this work addresses several challenges to existing FL algorithms, we overview the related work in three sub-topics of FL: works considering personalisation, works dealing with practical and deployment challenges, and works aiming to improve convergence speed and global-model performance.

\subsection{Personalised Federated Learning}
\textcolor{black}{Several authors have considered the approach of `personalising' FL models in order to tailor model performance to non-IID user datasets.}

\textcolor{black}{Meta-Learning aims to train a model that is easy to fine-tune with few samples. Fallah \textit{et al}. \cite{PerFLMAML} proposed the Per-FedAvg algorithm based on Model Agnostic Meta-Learning (MAML), that adds a  first-order adaptation term to the client loss functions, so they can be tuned to client datasets with one step. Jiang \textit{et al}. \cite{ImproveFLMAML} highlighted the connection between FedAvg and first-order MAML updates, and proposed a three-stage training algorithm to improve personalisation. }

\textcolor{black}{Other authors propose training a combination of local and global models in FL to improve personalisation. Hanzely and Richt{\'a}rick \cite{FedLearnMixture} added a learnable parameter to allow clients to control the extent of local and global model mixing. Dinh \textit{et al}. \cite{Moreau} kept a global model and a personal model for each user, performing SGD on their personal model and then updating their copy of the global model in an outer loop. Huang \textit{et al}. \cite{PerCrossSiloFL} kept a local model on each client, and added a proximal term to client loss functions to keep these models close to a `personalised' cloud model, for the cross-silo FL setting.}

\textcolor{black}{Smith \textit{et al}. \cite{FedMultiTaskLearn} proposed MOCHA, which performs Federated MTL formulates FL as a function of the model weight matrix and a relationship matrix. Their algorithm takes into account the heterogeneous hardware of clients, meaning MOCHA is not directly comparable to our MTFL scheme. Recently, Dinh \textit{et al}. \cite{FedU} generalised MOCHA and other algorithms into the FedU framework, including proposing a decentralised version.}

\textcolor{black}{Our work proposes a Multi-Task learning approach to achieve personalisation in FL (MTFL). We later show that our approach has substantial converge speed, personalisation performance, privacy and storage coast benefits compared to existing personalised FL algorithms. }

\subsection{Federated Learning in Edge Computing}
FL performs distributed computing at the network edge. Some authors have considered the system design and communication costs of FL in this environment. Jiang \textit{et al}. \cite{DistDLSystem} proposed an FL system that reduces the total data clients upload by selecting model weights with the largest gradient magnitudes. They also considered implementation details such as asynchronous or round-robin client updates. Bonawitz \textit{et al}. \cite{FedAvgSysDesign} produced a FedAvg system design, specifying  clients/server roles, fault handling, and security. They also provide analytics for their deployment of this system with over 10 million clients. To address the non-IID nature of client datasets in FL, Duan \textit{et al}. \cite{Astrea} proposed the Astrea framework: client datasets are augmented to help reduce local class imbalances, and mediators are introduced to the global aggregation method.

Several authors have also investigated the impacts of wirelessly connected FL clients. Yang \textit{et al}. \cite{SchedPolsFL} studied different scheduling policies in a wireless FL scenario. Their analysis showed that with a low Signal-to-Interference-plus-Noise Ratio (SINR), simple FL schemes perform well, but that as SINR increases, more intelligent methods of selecting clients are needed. Ahn \textit{et al}. \cite{WireFedDistill} proposed a Hybrid Federated Distillation scheme for FL with wireless edge devices, including using over-the-air computing and compression methods. Their results showed that their scheme gave better performance in high-noise wireless scenarios. 

Other authors have proposed schemes for considering the computing, networking and communication resources of FL clients in edge computing. Wang \textit{et al}. \cite{EffSchedMobDevices} performed experiments with smartphones to argue that the computation-time (as opposed to communication-time) of FedAvg is the most significant bottleneck for real-world FL, and propose algorithms to accommodate this computational heterogeneity. Nishio and Yonetani \cite{ClientSelectFL} designed a system that collects information about the computing and wireless resources of clients before initiating a round of FL, reducing the real-time taken to reach a target accuracy for FedAvg. 

These previous works have proposed implementations of FL systems. However, they do not consider MTL within FL, which is a main contribution of our work with the MTFL algorithm.

\subsection{Federated Learning Performance}
The seminal FedAvg algorithm \cite{FedAvgPaper} collaboratively trains a model by sending an initial model to participating clients, who each perform SGD on the model using their local data. These new models are sent to the server for averaging and a new round is begun. Some progress has been made towards improving the convergence rate of FedAvg. Leroy \textit{et al}. \cite{SNIPSPaper} used  Adam adaptive optimisation when updating the global model on the server. Reddi \textit{et al}. \cite{AdaptiveFedOpt} also generalised other adaptive optimisation techniques in the same style and provided convergence guarantees. Our FedAvg-Adam algorithm differs from these as clients in FedAvg-Adam perform Adam SGD (as opposed to vanilla SGD), and the Adam parameters are averaged alongside model weights at the server.  Liu \textit{et al}. \cite{AccFLMomentum} used momentum-SGD on clients, and aggregated the momentum values of clients on the server alongside the model weights as an alternative method of accelerating convergence.

Some works have been produced investigating FL with non-IID or poor-quality client data. Zhao \textit{et al}. \cite{FedLearnNonIID} proposed sharing a small amount of data between clients to decrease the differences in their data distributions and improve global model accuracy. Konstantinov and Lampert \cite{RobustLearningUntrusted} evaluated which clients have poor-quality data by finding the difference between a client model's local predictions and predictions using a trusted dataset. Wang \textit{et al}. \cite{CMFL} ignored \textit{irrelevant} client updates during training  by checking if each client's update aligns with the global model.

The FedAvg-Adam optimisation method presented here uses adaptive optimisation on clients, rather than SGD, which we later show  converges much faster than when using FedAvg or Adam optimisation purely on the server.

\section{Multi-Task Federated Learning (MTFL)}
Fig. 1 shows a high-level overview of how the MTFL algorithm would operate in the edge-computing environment. More detailed descriptions of the use of BN patches in MTFL, and optimisation on clients is given in the later subsections.

The MTFL algorithm is based on the client-server framework, however, rounds are initiated by the server, as shown in Fig 1. First, the server selects all, or a subset of all, known clients from its database and asks them to participate in the FL round (\textbf{Step 1}), and sends a \textit{Work Request} message to them. Clients will accept a \textit{Work Request} depending on user preferences (for example, users can set their device to only participate in FL if charging and connected to WiFi). All accepting clients then send an \textit{Accept} message to the server (\textbf{Step 2}). The server sends the global model (and any associated optimization parameters) to all accepting clients, who augment their copy of the global model with private patches (\textbf{Step 3}). Clients then perform local training using their own data, creating a different model. Clients save the patch layers from their new model locally, and upload their non-private model parameters to the server (\textbf{Step 4}). 

The server waits for clients to finish training and upload their models (\textbf{Step 5}). It can either wait for a maximum time limit, or for a given fraction of clients to upload before continuing, depending on the server preferences. After this, the server will aggregate all received models to produce a single global model (\textbf{Step 6}) which is saved on the server, before starting a new round.

MTFL therefore offloads the vast majority of computation to client devices, who perform the actual model training. It preserves users' data-privacy more strongly than FedAvg and other personalised-FL algorithms: not only is user data not uploaded, but key parts of their local models are not uploaded. The framework also accounts for client stragglers with its round time/uploading client fraction limit. Moreover, MTFL utilises patch layers to improve \textit{local} model performance on individual users' non-IID datasets, making MTFL more personalised.

\subsection{User Model Accuracy and MTFL}
In many FL works, such as the original FedAvg paper \cite{FedAvgPaper}, the authors use a central IID test-set to measure FL performance. Depending on the FL scenario, this metric may or may not be desirable. If the intention is to create a single model that has good performance on IID data, then this method would be suitable. However, in many FL scenarios, the desire is to create a model that has good performance on individual user devices. For example, Google have used FedAvg for their GBoard next-word-prediction software \cite{GBoard}. The objective was to improve the prediction score for individual users. As users do not typically have non-IID data, a single global model may display good performance for some users, and worse performance for others.

\begin{figure}[h]
\centering
\includegraphics[width=0.85\linewidth]{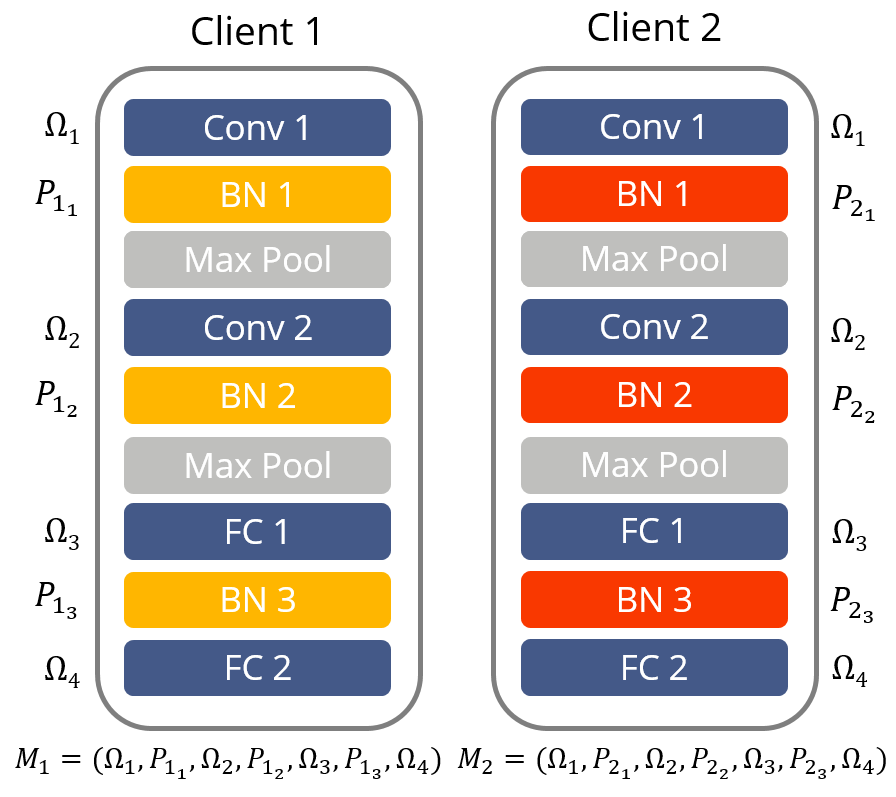}
\caption{Example composition of a DNN model used in MTFL. Each client's model consists of shared global parameters ($\Omega_1 - \Omega_4$) for Convolutional (Conv) and Fully-Connected (FC) layers, and private Batch-Normalization (BN) patch layers ($P_{k_1}, P_{k_2}, P_{k_3}$).}
\label{fig2}
\end{figure}

We propose using the average User model Accuracy (UA) as an alternative metric of FL performance. UA is the accuracy on a client using a local test-set. This test-set for each client should be drawn from a similar distribution as its training data. In this paper, we perform experiments on classification problems, but UA could be altered for different metrics (e.g. error, recall).

In FL, user data is often non-IID, so users could be considered as having different but related learning tasks. It is possible for an FL scheme to achieve good global-model accuracy, but poor UA, as the aggregate model may perform poorly on some clients' datasets (especially if they have a small number of local samples, so are weighted less in the FedAvg averaging step). We propose the MTFL algorithm that allows clients to build different models, while still benefiting from FL, in order to improve the average UA. Mudrakarta \textit{et al}. \cite{KFor1} have previously shown that adding small per-task `patch' layers to DNNs improved their performance in MTL scenarios. Patches are therefore a good candidate for training personalised models for clients. 

In FL, the aim is to minimise the following objective function:

\begin{equation}
F_{\mathrm{FL}} = \sum_{k=1}^K \frac{n_k}{n} \ell_k(\Omega) \\
\end{equation}

\noindent where $K$ is the total number of clients, $n_k$ is the number of samples on client $k$, $n$ is the total number of samples across all clients, $\ell_k$ is the loss function on client $k$, and $\Omega$ is the set of global model parameters. Adding unique client patches to the FL model changes the objective function of MTFL to:

\begin{equation}
F_{\mathrm{MTFL}} = \sum_{k=1}^K \frac{n_k}{n} \ \ell_k(\mathcal{M}_k) \\
\end{equation}
\begin{equation}
\mathcal{M}_k = (\Omega_1 \cdots \Omega_{i_1}, P_{k_1}, \Omega_{i_1 + 1} \cdots \Omega_{i_m} , P_{k_m}, \Omega_{i_m + 1} \cdots \Omega_j)
\end{equation}

\noindent \textcolor{black}{ where $\mathcal{M}_k$ is the patched model on client $k$, composed of Federated model parameters $\Omega_1 \cdots \Omega_j$ ($j$ being the total number of Federated layers) and patch parameters $P_{k_1} \cdots P_{k_m}$ ($m$ being the total number of local patches, $\{i\}$ being the set of indexes of the patch parameters) unique to client $k$. Fig. 2 shows an example composition of a DNN model used in MTFL.}

MTFL is a general algorithm for incorporating MTL into FL. Different optimisation strategies (including FedAvg-Adam described in Section 3.3) can be used within MTFL, and we later show that MTFL can substantially reduce the number of rounds to reach target UA, regardless of the optimisation strategy used.

\begin{algorithm}[h]
\caption{MTFL}
\begin{algorithmic}[1]

\State Initialise global model $\Omega$ and global optimiser values $V$
\While{termination criteria not met}
	\State Select round clients, $S_r \subset S$, $|S_r| = C \cdot |S|$
	\Statex 
	\ForEach{client $s_k \in S_r$ in parallel}
		\State Download global parameters $\mathcal{M}_k \gets \Omega$
		\State Download optimiser values $V_k \gets V$
		
		\For{$i \in$ patchIdxs} \Comment Apply local patches
			\State $\mathcal{M}_{k,i} \gets P_{k,i}$, $V_{k,i} \gets W_{k,i}$
		\EndFor		
		
		\For{batch $b$ drawn from local data $D_k$}
			\State $\mathcal{M}_k, V_k \gets \mathrm{LocalUpdate}(\mathcal{M}_k, V_k, b)$
		\EndFor
				
		\For{$i \in$ patchIdxs}	\Comment Save local patches
			\State $P_{k,i} \gets \mathcal{M}_{k,i}, W_{k,i} \gets V_{k,i}$
		\EndFor
				
		\ForEach{$i \notin$ patchIdxs}
			\State Upload $\mathcal{M}_{k,i}, V_{k,i}$ to server
		\EndFor	
	\EndFor
	
	\Statex
	\For{$i \notin $ nonPatchIndexes}
		\State $\Omega_i \gets \mathrm{GlobalModelUpdate}(\Omega_i, \{\mathcal{M}_{k,i} \}_{k \in S_r})$
		\State $V_i \gets \mathrm{GlobalOptimUpdate}(V_i, \{ V_{k,i} \}_{k \in S_r})$
	\EndFor 
\EndWhile
\end{algorithmic}
\end{algorithm}

As shown in Algorithm 1, MTFL runs rounds of communication until a given termination criteria (such as target UA) is met (Line 2). At each round, a subset $S_r$ of clients are selected to participate from the set of all clients $S$ (Line 3). These clients download the global model $\Omega$, which is a tuple of model parameters, and the global optimiser $V$, if used (Lines 5-6). The clients then update their copy of the global model and optimiser with their private patch layers (Lines 7-9), where the `patchIdxs' variable contains the indexes of patch layer placement in the DNN. Clients perform training using their now-personalised copy of the global model and optimiser on their local data (Line 10). Depending on the choice of FL optimisation strategy to be used within MTFL, the LocalUpdate function represents local training of the model. For FedAvg, LocalUpdate is simply minibatch-SGD. We discuss this, and the proposed FedAvg-Adam optimisation strategy, further in Section 3.3.  After local training, the updated local patches are saved (Lines 11-13), and the non-patch layers and optimiser values are uploaded to the server (Lines 14-16).

At the end of the round, the server makes a new global model and optimiser according to the GlobalModelUpdate and GlobalOptimUpdate functions (Lines 18-20). These functions are again dependent on the FL optimisation strategy used, and are discussed further in Section 3.3. FedAvg, for example, uses a weighted average of client models for GlobalModelUpdate. The updated global model marks the end of the round and a new round is begun.

\textcolor{black}{The total per-round computation complexity of MTFL scales with $|S_r|$, where $|S_r|$ is the number of clients participating per round. The computation performed by each client is independent of the total number of clients. As clients perform local computation in parallel, MTFL (like FedAvg) is eminently scalable. Scalability is important in FL as real-world deployments are expected to have huge numbers of low-powered clients \cite{FedLearnChallenges} \cite{GBoard}. The global model and optimiser updates (Lines 20-23 in Algorithm 1) depend on the optimisation strategy used. For FedAvg and FedAvg-Adam, GlobalModelUpdate is essentially map-reduce (averaging after local training) - also $\mathcal{O}(|S_r|)$. For FedAdam, the Adam step following the map-reduce in GlobalOptimUpdate is not dependent on the number of clients (only on the DNN architecture).}

\textcolor{black}{There are numerous works investigating FL in the Peer-To-Peer (p2p) setting, which we do not consider in this paper. Simple p2p FL algorithms involve sending all client models to all participating peers for decentralised aggregation. Extension of MTFL to these schemes is trivial: peers would simply just send/aggregate the non-private layers. More sophisticated p2p FL algorithms may require more complex ways of incorporating private layers – an interesting direction we leave for future works.}

Mudrakarta \textit{et al}. \cite{KFor1} showed that Batch Normalisation (BN) layers can act as model patches for MTL in the centralised setting. We show later that BN layers work well as  patches in MTFL, considering that they are very lightweight in terms of number of parameters. BN layers are given by:

\begin{equation}
\begin{gathered}
\hat{x}_i = \frac{z_i - \E(z_i)}{\sqrt{\mathrm{Var}(z_i) + \epsilon}} \\
\mathrm{BN}(\hat{x}_i) = \gamma_i \hat{x}_i + \beta_i
\end{gathered}
\end{equation}

\noindent where $\E(z_i)$ and $\mathrm{Var}(z_i)$ are the mean and variance of a neuron's activations ($z_i$, post-nonlinearity) across a minibatch, and  $\gamma_i$ and $\beta_i$ are parameters learned during training. BN layers track a weighted moving average of $\E(z_i)$ and $\mathrm{Var}(z_i)$ during training: $\mu_i$ and $\sigma_i^2$, for use at inference time. In Section 4 we investigate the benefit of keeping statistics $\mu, \sigma$ and/or trainable parameters $\gamma, \beta$ as part of private patch layers. 

\textcolor{black}{We have chosen to use BN layers for personalisation within MTFL. The reason for this choice is twofold: 1) they show excellent personalisation performance and 2) the storage cost of BN parameters is very small ($<1\%$ of total model size for the tested model architectures). Mudrakarta \textit{et al}. \cite{KFor1} also investigated the use of depthwise-convolutional patches for centralised Multi-Task learning. Any model layers could in principle be kept private during MTFL, however, there is an inherent trade-off between the number of parameters kept private and the ability of the global model to converge.}

\subsection{Effect of BN Patches on Inference}
\begin{figure}[h]
\centering
\includegraphics[width=0.8\linewidth]{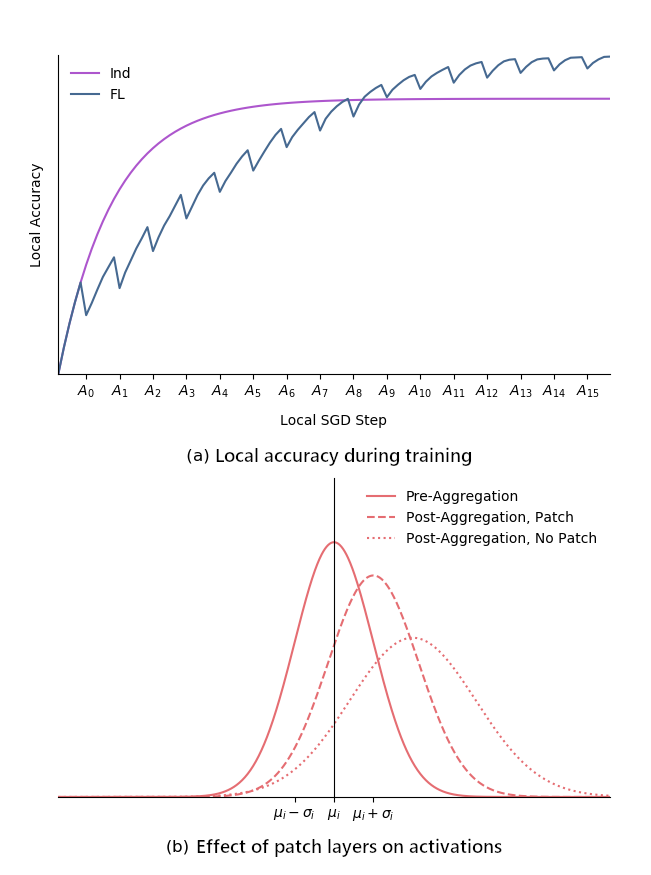}
\caption{(a) Federated Learning (FL) results in an accuracy curve where the UA decreases after aggregations and increases during local training, compared with the smoother accuracy curve when training independently (Ind). (b) Patch BN-layers help bring the distribution in outputs for neuron $i$ closer to the pre-aggregation distributions.}
\label{fig2}
\end{figure}

\noindent To understand the impact that BN-patch layers have on UA, we consider the change in internal DNN activations over a client's local test-set immediately \textit{before} and immediately \textit{after} the FL aggregation step. 

As illustrated in Fig. 3 (a), UA typically drops after the aggregation step in iterative FL. This is because the model has been tuned on the local training set for several epochs, and suddenly has its model weights replaced by the Federated weights, which are unlikely to have better test performance than the pre-aggregation model. This idea is further examined in \cite{SalvagingFL} and showed later in our experimental section. Consider a simple DNN consisting of dense layers followed by BN and then nonlinearities. The vector of first-layer neuron activations over the client's test-set ($X$) from applying weights and biases ($W_0$, $b_0$), can be modelled as a normal distribution, which BN relies on to work:

\begin{equation}
\begin{gathered}
\textcolor{black}{z_i \triangleq \lbrack W_0 X + b_0 \rbrack_i }\\ 
\textcolor{black}{z_i \sim N(\E[z_i], Var[z_i]) }
\end{gathered}
\end{equation}

\noindent During local training, the client's model has been adapted to the local dataset, and the BN-layer statistics used for inference ($\mu$, $\sigma^2$) have been updated from the layer activations. Assuming, after local training (and before aggregation), $\mu_i \approx \E[z_i]$, and $\sigma_i^2 \approx Var[z_i]$, then the BN-layer (ignoring $\epsilon$) computes:

\begin{equation}
\begin{gathered}
\textcolor{black}{\hat{x}_i \triangleq \frac{z_i - \mu_i}{\sigma_i}} \\
\textcolor{black}{\hat{x}_i \sim N(0, 1)} \\
\textcolor{black}{\mathrm{BN}(\hat{x}_i) \sim N(\beta_i, \gamma_i^2)}
\end{gathered}
\end{equation}

\noindent where $\beta_i$, $\gamma_i^2$ are the learned BN parameters. If the client is participating in  FL or MTFL, then the model parameters $W_0$, $b_0$ are updated after downloading the global model with federated values: $\overline{W}_0$, $\overline{b}_0$. The activations of the first layer are then:

\begin{equation}
\begin{gathered}
\textcolor{black}{\overline{z}_i \triangleq \lbrack \overline{W}_0 X + \overline{b}_0 \rbrack_i} \\ 
\textcolor{black}{\overline{z}_i \sim N(\E[\overline{z}_i], \mathrm{Var}[\overline{z}_i])}
\end{gathered}
\end{equation}

\noindent Defining the difference in mean and variance between pre- and post-aggregation activations, $\Delta \mu_i = \E[\overline{z}_i] - \E[z_i]$ and $\Delta \sigma_i^2 = \mathrm{Var}[\overline{z}_i] - \mathrm{Var}[z_i]$, the output from a BN-patch layer as part of MTFL (which maintains $\mu$, $\sigma$, $\beta$, $\gamma$ after aggregation) is:

\begin{equation}
\begin{gathered}
\hat{\bar{x}}_i \sim N \left( \frac{\Delta \mu_i}{\sigma_i}, 
						      1 + \frac{\Delta \sigma_i^2}{\sigma_i^2} \right) \\
\mathrm{BN}(\hat{\bar{x}}_i) \sim N \left( \gamma \frac{\Delta \mu_i}{\sigma_i} + \beta_i, 
                                     \gamma_i^2 \left( 1 + \frac{\Delta \sigma_i^2}{\sigma_i^2} \right) \right)
\end{gathered}
\end{equation}

\noindent If the BN layer is \emph{not} a patch layer (i.e., the client is participating in FL, with federated BN values $\bar{\mu}$, $\bar{\sigma}$, $\bar{\beta}$, $\bar{\gamma}$), the output of the BN layer is:

\begin{equation}
\begin{gathered}
\hat{\bar{x}}_i \sim N \left( 	\frac{\mu_i+\Delta\mu_i-\bar{\mu}_i}{\bar{\sigma}_i}, 
								\frac{\sigma_i^2+\Delta\sigma_i^2}{\bar{\sigma}_i^2} \right) \\
\overline{\mathrm{BN}}(\hat{\bar{x}}_i) \sim N \left( 
			\bar{\gamma} \frac{\mu_i+\Delta\mu_i-\bar{\mu}_i}{\bar{\sigma}_i} + \bar{\beta}_i,
		 	\bar{\gamma}_i^2 \frac{\sigma_i^2+\Delta\sigma_i^2}{\bar{\sigma}_i^2} \right)
\end{gathered}
\end{equation}

\noindent We posit that using BN-patch layers in MTFL constrains neuron activations to be closer to what they were before the aggregation step, compared to non-patch BN layers as part of FL (as illustrated in Fig. 3 (b). I.e., the difference in means and variances pre- and post-aggregation using MTFL is smaller than when using FL:

\begin{equation}
\begin{gathered}
\left| \gamma\frac{\Delta\mu_i}{\sigma_i} \right| < 
		\left| \beta_i - \bar{\gamma} \frac{\mu_i+\Delta\mu_i-\bar{\mu}_i}{\bar{\sigma}_i} - \bar{\beta}_i \right| \\ 
\left| \gamma_i^2\frac{\Delta \sigma_i^2}{\sigma_i^2} \right| < 
	\left| \gamma_i^2 - \bar{\gamma}_i^2 \frac{\sigma_i^2+\Delta\sigma_i^2}{\bar{\sigma}_i^2} \right|
\end{gathered}
\end{equation}

\noindent Assuming the above inequality holds, it is easy to see how the values propagated through the network after the first layer are closer to the pre-aggregation values when using BN-patches as opposed to federated BN layers. If BN-patches are added throughout the network, the intermediate DNN values will be regularly `constrained' to be closer to the pre-aggregation values, resulting ultimately in network outputs closer to the pre-aggregation outputs.

Looking at Eq. (8), if $\Delta \mu_i$ and $\Delta \sigma_i^2$ for neuron $i$ are large, then the output distribution of the neuron after the BN-patch layer ($\mathrm{BN}(\hat{\bar{x}})_i$) over the test-set will be quite different than $\mathrm{BN}(\hat{x})_i$. The BN-patch layer will therefore provide little benefit over a federated BN layer, as the left hand sides of the inequalities in Eq. (10) are unlikely to be much smaller than the right hand sides. Large differences in pre- and post-aggregated model parameters are seen during the early stages of training, when gradients are large and client models diverge more during local training. This therefore implies that MTFL has less benefit during the early stages of training, and its benefit increases during training as gradient magnitudes decrease (as shown in Fig. 4).

\subsection{Federated Optimisation within MTFL}
As shown in Algorithm 1, MTFL applies private patch layers for each client, and trains them alongside the federated (non-private) layers during LocalUpdate. At the end of each round, the server aggregates the uploaded federated layers from clients (and any distributed optimiser values used), producing a new global model using the GlobalModelUpdate function. If distributed adaptive-optimisation is used, then the GlobalOptimUpdate function will also be called. Table 1 details different FL training algorithms as characterised by their implementations of these functions. 

In FedAvg, LocalUpdate is simply minibatch-SGD, and GlobalModelUpdate produces the new global model as a weighted (by number of local samples) average of uploaded client models. FedAvg uses SGD with no adaptive optimisation, so the variable $V$ in Algorithm 1 is a tuple of empty values, and GlobalOptimUpdate performs no function. For FedAdam \cite{SNIPSPaper} \cite{AdaptiveFedOpt}, clients also perform SGD during LocalUpdate. However, during GlobalModelUpdate, the server takes the difference ($\Delta_r$) between the previous global model and the average uploaded client model. The server treats $\Delta_r$ as a `psuedogradient', and uses a set of \nth{1} and \nth{2} moment values stored on the server to update the global model using an Adam-like update step. Clients do not use distributed adaptive optimisation in FedAdam, so $V$ is also a tuple of empty values and GlobalOptimUpdate performs no function.

We propose using adaptive optimisation (namely, Adam) as the distributed optimisation strategy. We call this strategy FedAvg-Adam. In FedAvg-Adam, clients share a global set of Adam \nth{1} and \nth{2} moments, stored in the $V$ variable in Algorithm 1. Clients store private optimiser values for their patch layers ($W_k$), as we find performance is better when keeping private optimiser values for patches. During LocalUpdate, clients perform Adam SGD, and the federated model layers and Adam values are uploaded by clients at the end of the round. To produce a new global model, the server averages the client models in GlobalModelUpdate and averages the Adam moments in GlobalOptimUpdate. FedAvg-Adam therefore has a $3\times$ communication cost per round compared to FedAvg or FedAdam. However, in many FL scenarios, the major concern is reducing the number of communication rounds required for the model to converge. We later show that FedAvg-Adam considerably improves the convergence speed of FL and MTFL.

For the rest of the paper, we refer to iterative FL schemes that do not keep any private model patches as FL, with the optimisation strategy in brackets, e.g. FL(FedAvg). If clients keep private model patches, we refer to the scheme as MTFL, again with the optimisation strategy in brackets, e.g. MTFL(FedAvg).

\begin{table}[t]
\centering
\caption{LocalUpdate, GlobalModelUpdate and GlobalOptimUpdate used by the FedAvg \cite{FedAvgPaper}, FedAdam \cite{SNIPSPaper} \cite{AdaptiveFedOpt} and FedAvg-Adam FL training strategies. All of these strategies can be used within MTFL.}
\small
\begin{tabular}{c|ccc}
	\toprule 
		Optimisation & \multirow{2}{*}{LocalUpdate} & GlobalModel & GlobalOptim \\
		Strategy & & Update & Update \\
	\midrule
		FedAvg & SGD & Average & - \\
		FedAdam & SGD & Adam & - \\
		FedAvg-Adam & Adam & Average & Average \\
	\bottomrule
\end{tabular}
\end{table}

\begin{table*}[t]
\centering
\caption{Communication rounds required to reach target average user accuracies for different tasks using FL and MTFL (with private statistics $\mu, \sigma$ and/or trained parameters $\gamma, \beta$), for different numbers of total clients $W$, client participation rates $C$, and optimisation strategies. Cases unable to reach the target UA within 500 rounds are denoted by X. Best results for each scenario (combination of $W$ and $C$) given in bold.}
\small
\begin{tabular}{ccccc|cccc|cccc|cccc}
	\toprule 
		\multicolumn{17}{c}{MNIST - 2NN}\\
	\midrule
		& \multicolumn{4}{c|}{\textbf{FL}} & \multicolumn{12}{c}{\textbf{MTFL}} \\
	     & \multicolumn{4}{c|}{Private values = None} & \multicolumn{4}{c|}{$\mu, \sigma, \gamma, \beta$} & \multicolumn{4}{c|}{$\mu, \sigma$} & \multicolumn{4}{c}{$\gamma, \beta$} \\
		 Optimisation & \multicolumn{2}{c}{W = 200} & \multicolumn{2}{c|}{400} & \multicolumn{2}{c}{200} & \multicolumn{2}{c|}{400} & \multicolumn{2}{c}{200} & \multicolumn{2}{c|}{400} & \multicolumn{2}{c}{200} & \multicolumn{2}{c}{400} \\
		 Strategy & C = 0.5 & 1.0 & 0.5 & 1.0 & 0.5 & 1.0 & 0.5 & 1.0 & 0.5 & 1.0 & 0.5 & 1.0 & 0.5 & 1.0 & 0.5 & 1.0\\
	\midrule
		FedAvg & \CB 99 & \CB 102 & \CB 107 & \CB 110 & \CB 85 & \CB 58 & \CB 101 & \CB 68 & \CB X & \CB X & \CB X & \CB X & \CB 29 & \CB 21 & \CB 34 & \CB 26 \\
		FedAdam & \CO 85 & \CO 69 & \CO 88 & \CO 65 & \CO 56 & \CO 37 & \CO 75 & \CO 77 & \CO 109 & \CO 90 & \CO 194 & \CO 262 & \CO 31 & \CO 25 & \CO 31 & \CO 27 \\
		FedAvg-Adam & \CG 44 & \CG 49 & \CG 40 & \CG 50 & \CG 17 & \CG 41 & \CG 19 & \CG 32 & \CG 131 & \CG 151 & \CG 170 & \CG 198 & \CG \textbf{9} & \CG \textbf{9} & \CG \textbf{10} & \CG \textbf{9} \\
	\midrule
		\multicolumn{17}{c}{CIFAR10 - CNN}\\
	\midrule
		FedAvg & \CB 139 & \CB 138 & \CB 171 & \CB 164 & \CB 49 & \CB 33 & \CB 55 & \CB 36 & \CB 231 & \CB 280 & \CB 258 & \CB 266 & \CB 37 & \CB 24 & \CB 45 & \CB 30 \\
		FedAdam & \CO 105 & \CO 90 & \CO 83 & \CO 80 & \CO 21 & \CO 14 & \CO 22 & \CO 16 & \CO 67 & \CO 45 & \CO 48 & \CO 38 & \CO 24 & \CO 14 & \CO 25 & \CO 16 \\
		FedAvg-Adam & \CG 57 & \CG 43 & \CG 36 & \CG 31 & \CG 11 & \CG 9 & \CG 14 & \CG \textbf{8} & \CG 82 & \CG 79 & \CG 62 & \CG 63 & \CG \textbf{10} & \CG \textbf{7} & \CG \textbf{11} & \CG \textbf{8} \\
	\bottomrule
\end{tabular}
\end{table*}

\begin{table*}[t]
\centering
\caption{Communication rounds required to reach target average user accuracies (of non-noisy clients) for different tasks using FL and MTFL (with private statistics $\mu, \sigma$ and/or trained parameters $\gamma, \beta$), when 20\% of clients have noisy training data, for different numbers of total clients $W$, client participation rates $C$, and optimisation strategies. Cases unable to reach the target UA within 500 rounds are denoted by X. Best results for each scenario (combination of $W$ and $C$) given in bold.}
\small
\begin{tabular}{ccccc|cccc|cccc|cccc}
	\toprule 
		\multicolumn{17}{c}{MNIST - 2NN}\\
	\midrule
	    & \multicolumn{4}{c|}{\textbf{FL}} & \multicolumn{12}{c}{\textbf{MTFL}} \\
	    & \multicolumn{4}{c|}{Private values = None} & \multicolumn{4}{c|}{$\mu, \sigma, \gamma, \beta$} & \multicolumn{4}{c|}{$\mu, \sigma$} & \multicolumn{4}{c}{$\gamma, \beta$} \\
		Optimisation & \multicolumn{2}{c}{W = 200} & \multicolumn{2}{c|}{400} & \multicolumn{2}{c}{200} & \multicolumn{2}{c|}{400} & \multicolumn{2}{c}{200} & \multicolumn{2}{c|}{400} & \multicolumn{2}{c}{200} & \multicolumn{2}{c}{400} \\
		Strategy & C = 0.5 & 1.0 & 0.5 & 1.0 & 0.5 & 1.0 & 0.5 & 1.0 & 0.5 & 1.0 & 0.5 & 1.0 & 0.5 & 1.0 & 0.5 & 1.0\\
	\midrule
		FedAvg & \CB 276 & \CB X & \CB 290 & \CB X & \CB 115 & \CB 76 & \CB 144 & \CB 144 & \CB 85 & \CB 58 & \CB 102 & \CB 68 & \CB 50 & \CB 36 & \CB 65 & \CB 48 \\
		FedAdam & \CO X & \CO X & \CO X & \CO X & \CO 76 & \CO 47 & \CO 110 & \CO 89 & \CO 56 & \CO 37 & \CO 75 & \CO 77 & \CO 43 & \CO 33 & \CO 46 & \CO 53 \\
		FedAvg-Adam & \CG 133 & \CG 260 & \CG 191 & \CG X & \CG 20 & \CG 16 & \CG 24 & \CG 27 & \CG 17 & \CG 41 & \CG 19 & \CG 32 & \CG \textbf{12} & \CG \textbf{8} & \CG \textbf{15} & \CG \textbf{40} \\
	\midrule
		\multicolumn{17}{c}{CIFAR10 - CNN}\\
	\midrule
		FedAvg & \CB 148 & \CB 208 & \CB 202 & \CB 250 & \CB 47 & \CB 32 & \CB 52 & \CB 35 & \CB 239 & \CB 186 & \CB 260 & \CB 88 & \CB 36 & \CB 24 & \CB 43 & \CB 28 \\
		FedAdam & \CO 159 & \CO 91 & \CO 92 & \CO 93 & \CO 21 & \CO 14 & \CO 21 & \CO 15 & \CO 74 & \CO 49 & \CO 51 & \CO 42 & \CO 34 & \CO 16 & \CO 21 & \CO 14 \\
		FedAvg-Adam & \CG 193 & \CG X & \CG X & \CG X & \CG 14 & \CG 10 & \CG 16 & \CG \textbf{9} & \CG 103 & \CG 111 & \CG 67 & \CG 74 & \CG \textbf{12} & \CG \textbf{8} & \CG \textbf{13} & \CG \textbf{9} \\
	\bottomrule
\end{tabular}
\end{table*}

\section{Experiments}
In this section, we first give details of the datasets, models and data partitioning scheme used for all the experiments. We then present extensive experiments analysing the impact that MTFL has on the number of rounds taken to reach a target UA. These experiments also examine which BN values, when kept private, give the best performance, and compare FL and MTFL with different optimisation strategies. After that, we investigate why different private BN values have different impacts on training, and compare MTFL to other state-of-the-art personalised FL algorithms. Finally, we evaluate the cost in terms of computation time of MTFL(FedAvg-Adam) on an MEC-like testbed. 

\subsection{Datasets and Models}
We conduct experiments using two image-classification datasets: MNIST \cite{MNIST} and CIFAR10 \cite{CIFAR}, and two DNN architectures. \\

\setlength{\leftskip}{0.75cm}
\setlength{\rightskip}{0.75cm}
\noindent
\textit{MNIST:} $(28 \times 28)$ pixel greyscale images of handwritten digits from 10 classes. The `2NN' network used on this dataset had one Fully Connected (FC) layer of 200 neurons, a BN layer, a second 200-neuron FC layer, and a softmax output layer. \\

\setlength{\leftskip}{0.75cm}
\setlength{\rightskip}{0.75cm}
\noindent
\textit{CIFAR10:} $(32 \times 32)$ pixel RGB images of objects from 10 classes. The `CNN' network used on this dataset had one $(3 \times 3)$ convolutional (conv) layer with 32 filters followed by BN, ReLU and $(2 \times 2)$ max pooling; a second $(3 \times 3)$ conv ReLU layer with 64 filters, BN, ReLU and $(2 \times 2)$ max pooling; a 512 neuron ReLU FC layer; and a softmax output layer. \\

\setlength{\leftskip}{0cm}
\setlength{\rightskip}{0cm}
\noindent Experiments were run with different numbers of clients $W$, client participation rates $C$ and optimisation strategies, on non-IID clients. To produce non-IID client data, we take the popular approach from \cite{FedAvgPaper}: order the training and testing data by label, split each into $2W$ shards, and assign each client two shards at random. Using the same assignment indexes for the testing data means that the classes in each client's training set are the same as those in their test set. This splitting produces a strongly non-IID distribution across clients. All results are an average over 5 trials with different random seeds.

\begin{figure*}[t]
\centering
\includegraphics[width=1.0\textwidth]{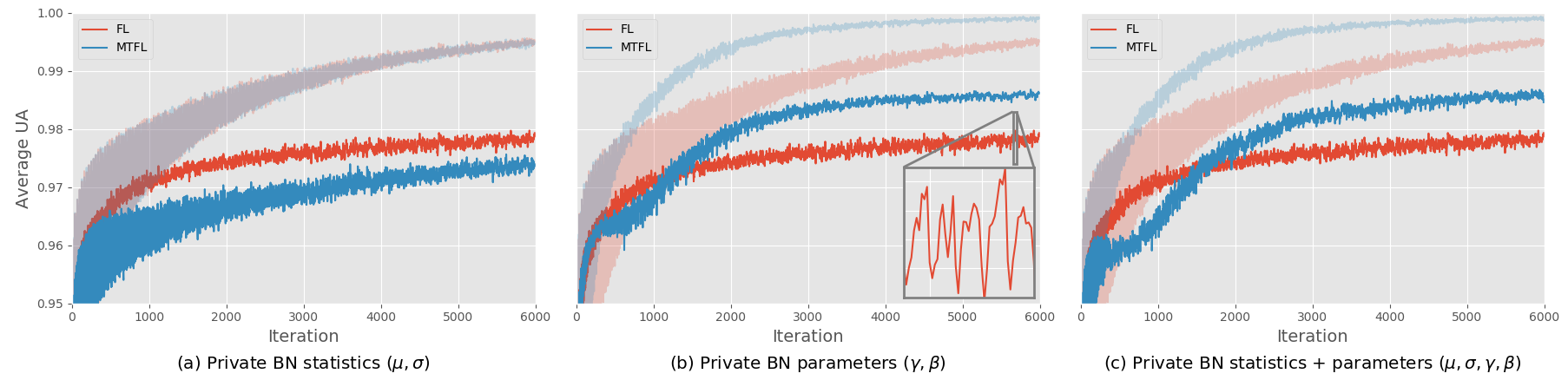}
\caption{Average training (faint) and testing (solid) User Accuracy (UA) curves for every step of local SGD on the MNIST, $W = 200$, $ C = 1.0$ scenario, using FL(FedAvg) (red), and MTFL(FedAvg) (blue). Each plot compares keeping different values within the BN layers of MTFL private: either statistics ($\mu, \sigma$) and/or trainable parameters ($\gamma, \beta$), to FL. All curves have been smoothed with an averaging kernel for presentation, except the inset of plot (b), which shows the cyclic drops in accuracy due to model averaging characteristic of FL.}
\label{fig2}
\end{figure*}

\subsection{Patch Layers in FL}
\textbf{Setup -} First, we compare how many rounds are needed to reach a target average UA (97\% for MNIST, 65\% for CIFAR10) for MTFL and FL. In FL, no model parameters are kept private (i.e. there are no patches), whereas in MTFL, some model parameters are kept private. For the MTFL columns in Tables 2 and 3, we present the effects of keeping BN-layer statistics ($\mu, \sigma$) and/or trainable parameters ($\gamma, \beta$) private, as part of the patch layers. 

For these results, we fixed number of local epochs $E = 1$ and tuned the learning-rate hyperparameters for every scenario such that the target was reached in the fewest rounds. For FedAvg and FedAvg-Adam, we had to tune only one hyperparameter for each scenario, but FedAdam required training both client and server learning rate. Entries with `X' in Tables 2 and 3 denote cases that could not reach the target within 500 communication rounds. 

In Table 3, we also investigate the robustness of MTFL to clients with noisy training data. Here, 20\% of the clients at random had 0-mean Gaussian noise added to their training data. The average UA taken for Table 3 was for the non-noisy clients only, to test how MTFL helps to mitigate the effect of noisy clients on non-noisy clients. We used Gaussian noise with standard deviation 3 for MNIST and 0.2 for CIFAR10 (MNIST is a much simpler image classification task than CIFAR10, so required more noise to significantly hinder training). \\

\noindent \textbf{Results -} Table 2 shows that for MTFL, when all BN-layer values ($\mu, \sigma, \gamma, \beta$) are kept private, the number of rounds to reach a target average UA is substantially lower in almost all scenarios when compared to FL. For example for the CIFAR10 $W = 400$, $C = 1.0$ scenario, FL(FedAvg) took 164 rounds to reach the target average UA, whereas MTFL(FedAvg) with private ($\mu, \sigma, \gamma, \beta$) took only 36 rounds. However, Tables 2 and 3 show that when keeping only the tracked statistics of BN patches private, UA is actually harmed. Conversely, MTFL with only private trainable parameters took even fewer rounds than MTFL will all-private ($\mu, \sigma, \gamma, \beta$).  For the same scenario, MTFL(FedAvg) with private ($\mu, \sigma$) took 266 rounds, whereas MTFL(FedAvg) with private ($\gamma, \beta$) took just 30 rounds. We investigate the reason behind these differences further in Section 4.3.

\textcolor{black}{MTFL naturally increases the variance of UAs during training, as non-identical user models in MTFL would contribute to the variance of UAs. However, in the experiments we performed, the difference between the variance of UA for FL and MTFL is very small: at less than 1\%.}

Table 3 shows that MTFL also helped to mitigate the impact of noisy clients on non-noisy clients. With FL, noisy clients prevented the average non-noisy UA from reaching the target in many scenarios. However, in most cases, MTFL allowed the non-noisy clients to reach the target average UA in a similar number of rounds than the corresponding non-noisy scenarios in Table 2. For example, for the CIFAR10 $W = 400, C=1.0$ scenario, FL(FedAvg) took 250 rounds to reach the target, however MTFL(FedAvg) with private ($\gamma, \beta$) parameters, took just 28 rounds.

\textcolor{black}{As Table 3 displays the rounds required for the non-noisy clients to reach the target average UA, the improvements shown when using MTFL may be due to the non-noisy clients being more `decoupled' from the noisy ones. As they do not share all model parameters, the harmful effect of receiving a global model that has been harmed by the participation of noisy clients has been reduced, allowing them to reach higher accuracies, faster.}

Tables 2 and 3 also show that in most scenarios, the FedAvg-Adam optimisation strategy reached the target average UA in the fewest rounds, regardless of whether FL or MTFL is used. Taking the same CIFAR10 scenario in Table 2, FL(FedAvg) took 164 rounds, FL(FedAdam) 80 rounds, and FL(FedAvg-Adam) only 31 rounds to reach the target. Similarly, MTFL(FedAvg) took 36 rounds, MTFL(FedAdam) 16 rounds and MTFL(FedAvg-Adam) just 8 rounds with private ($\mu, \sigma, \gamma, \beta$).

\begin{figure*}[h]
\centering
\includegraphics[width=1.0\textwidth]{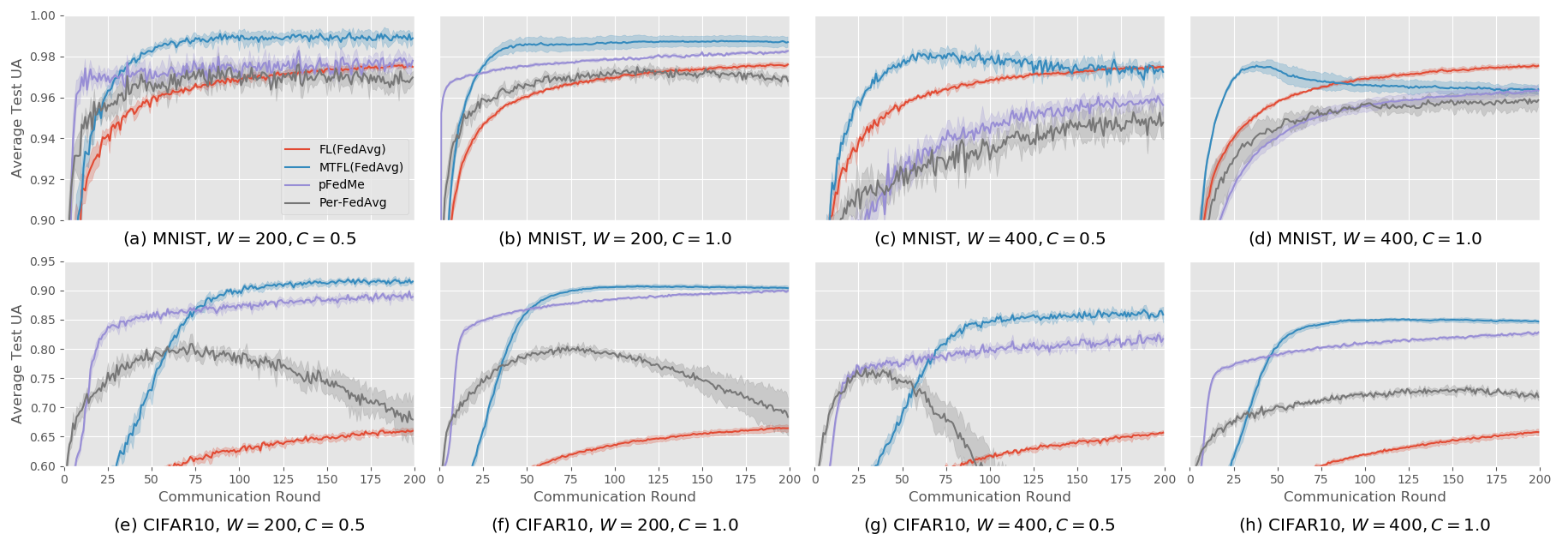}
\caption{Per-round testing User Accuracy (UA) of four FL algorithms: FL(FedAvg) \cite{FedAvgPaper}, MTFL(FedAvg) (using private $\gamma, \beta$), pFedMe \cite{Moreau} and Per-FedAvg \cite{PerFLMAML}. Experiments are conducted on MNIST and CIFAR10, with data divided in a non-IID fashion between $W = 200$ or $400$ clients, and $C = 0.5$ or $1.0$ fraction of users participating per round. Shaded regions show 95\% confidence intervals per round over 5 trials with different random seeds.}
\label{fig2}
\end{figure*}

\subsection{Training and Testing Results Using MTFL}
\textbf{Setup -} To investigate why MTFL with BN patches using private ($\mu, \sigma$) and/or private ($\gamma, \beta$) give such different results (as shown in Tables 2 and 3), we plotted training and testing UA during one scenario from Table 2: namely MNIST with $W = 200$, $C = 1.0$ for FL(FedAvg) and MTFL(FedAvg). We ran the algorithms for 600 communication rounds, where clients performed 10 steps of local training each round, and calculated the average training and testing UA for every local step. These graphs therefore present $600 \times 10 = 6000$ total steps. Measuring in this way allowed us to present the train/test accuracy trade-off, the impact that averaging has during training, and the effect on training and testing with different private BN values. \\

\noindent \textbf{Results -} Fig. 4 shows the (smoothed) training and testing UAs of the different combinations of BN layer statistics/parameters for the MNIST problem. Note that because the lines are smoothed for presentation, the steps where the curves reach the target accuracies may not correspond to the values in Table 2. In Fig. 4 (a), the training curves for FL(FedAvg) and MTFL(FedAvg) with private ($\mu, \sigma$) are the same: this is because the BN statistics are only used at test-time and do not influence training. The test accuracy for private ($\mu, \sigma$) is lower than FedAvg, mirroring the results in Tables 2 and 3. The lower test accuracy may be due to mismatch in BN values: $\gamma$ and $\beta$ have been averaged, so output a different distribution than these private statistics have been tracking, thus harming the ability of the model. This seems to be supported by Fig. 4 (c). When keeping both private ($\mu, \sigma$) and ($\gamma, \beta$) there is no substantial performance drop when compared to Fig 4 (b), when only ($\gamma, \beta$) are kept private. 

Fig 4. (b) and (c) show that keeping private ($\mu, \sigma$) significantly increases the rate at which the training accuracy can improve (see faint lines in Fig 4. (b) and (c)). Previous authors \cite{FedAvgPaper} have commented that FedAvg can work as a kind of regularisation for client models. When clients have small local datasets, their training error would quickly reach near-0 as it is easy for independently-trained models to overfit. However, they would have poor generalisation performance (which is the motivation behind FL). Keeping some model parameters private (here $\mu$ and $\sigma$ from the BN layers) seems to strike a balance between fast convergence (which would be achieved by a fully-private model) and regularisation due to FL (which is achieved by averaging client model parameters).

\subsection{Personalised FL Comparison}
\textbf{Setup -} We compare the personalisation performance of MTFL(FedAvg) with two other state-of-the-art Personalised FL algorithms: Per-FedAvg \cite{PerFLMAML} and pFedMe \cite{Moreau}, and FL(FedAvg) (where no model layers are private) \cite{FedAvgPaper}. We tuned the hyperparameters of each algorithm to achieve the maximum average UA within 200 communication rounds. We present MTFL(FedAvg) with private ($\gamma, \beta$), not MTFL(FedAvg-Adam), as we wish only to compare the personalisation algorithms, not the benefit of the adaptive optimisation strategy. We also fixed the amount of local computation to be roughly constant for the algorithms: we perform $E = 1$ epoch of local training for MTFL(FedAvg) and FL(FedAvg). For MNIST, using a batch size of 20, this is equivalent to 15 and 8 steps of local SGD for $W = 200$ and $W = 400$, respectively. For CIFAR10, this is equivalent to 13 and 7 steps of local SGD for $W = 200$ and $W = 400$, respectively. Per-FedAvg uses the value $K$ for local iterations, so we fix that the the same number of steps for FL and MTFL. pFedMe uses has two inner loops, and we set the number of outer-loops $R$ to the same value as $K$ from Per-FedAvg, and fix the inner-loop number for pFedMe to 1 for all scenarios. This setup results in the same number of local steps performed for each algorithm, however, the cost per local step of Per-FedAvg and pFedMe is considerably higher than FL(FedAvg) and MTFL(FedAvg). Note also that MTFL(FedAvg) and FL(FedAvg) have only one hyperparameter, $\eta$ to tune, whereas Per-FedAvg and pFedMe both have two. This makes the hyperparameter search for Per-FedAvg and pFedMe considerably more costly. \\

\noindent \textbf{Results -} The plots in Fig. 5 show that MTFL(FedAvg) was able to achieve a higher UA compared to the other schemes in all tested scenarios. Per-FedAvg and pFedMe were able to reach a higher UA than FL(FedAvg) in the $W = 200$ cases for MNIST, but were actually slower in the $ W = 400$ cases. All the personalised-FL schemes were able to achieve good UA faster than FL(FedAvg) for the CIFAR10 experiments, however. This is likely due to the CIFAR10 task being a much harder one than MNIST. It is interesting to note that Per-FedAvg appeared to overfit quickly on this task. Also worthy of note is the fact that MTFL(FedAvg) was able to beat Per-FedAvg and pFedMe whilst also having one less hyperparameter to tune, and being computationally cheaper. MTFL also provides the extra benefit to privacy of keeping some model parameters private (pFedMe and Per-FedAvg both upload entire models).

\subsection{Testbed Results}
\textbf{Setup -} To test MTFL in a more realistic MEC environment, we set up a testbed consisting of 10 clients: 5 Raspberry Pi (RPi) 2B's and 5 RPi 3B's, connected over WiFi to a server, in order to emulate a low-powered, heterogeneous set of clients. The RPi's used Tensorflow to perform local training. The server did not perform any model testing, only receiving, averaging and distributing models. The average time over 10 rounds was taken, along with the percentage of time spent per round in downloading models from the server, local training, uploading models and work performed on the server. \\

\noindent \textbf{Results -} Table 4 shows the average time taken per round for FL(FedAvg), MTFL(FedAvg-Adam), and Independent learning, when one local epoch of training is performed. Each round is also split by time spent for each task within the round. As would be expected, Independent learning took the least time per round as clients did not have to download / upload any models. FL(FedAvg) took longer per round due to uploading / downloading, and MTFL(FedAvg-Adam) took the longest per round due to the increased number of weights that FedAvg-Adam communicates over FedAvg, indicated by the higher percentage of round time spent downloading and uploading models. However, the increase in communication time is likely to be outweighed in most cases by the far fewer rounds required to reach a target average UA (see Tables 2-3).

The majority of the round times were spent in local training rather than in communication for FL or MTFL. This is due to the low computing power of the RPi's and the high computational cost of training DNN models. In real-world FL scenarios, the round times are influenced by the compute abilities of client devices, the computational cost of the models used, and the communication conditions.

\addtolength{\tabcolsep}{-1.85pt}    
\begin{table}[h]
\centering
\caption{Average time per round of different learning schemes on the MNIST and CIFAR10 datasets, and percentage of time spent downloading the model (\textit{Down}), training the model (\textit{Client}), uploading the model (\textit{Up}), and model aggregation/distribution on the server (\textit{Server}) took.}
\small 
\begin{tabular}{cccccc}
\toprule 
\MCR{6}{1}{MNIST - 2NN} \\
Learning    & Round     & \MCR{4}{1}{Percentage of Round Time (\%)} \\ 
Scheme      & Time (s)  &\textit{Down} & \textit{Client} & \textit{Up} & \textit{Server} \\ 
\midrule 
FL(FedAvg)		& 30 	    & 5 	& 88 	& 6 	& 1 	        \\
MTFL(FedAvg-Adam)	& 38 	    & 11 	& 76 	& 12 	& 1 	        \\
Independent & 29 	    & 0 	& 100 	& 0 	& 0 	        \\
\midrule 
\MCR{6}{1}{CIFAR10 - CNN} \\ 
\midrule 
FL(FedAvg) 		& 108 	    & 5 	& 86 	& 5 	& 4 	        \\
MTFL(FedAvg-Adam)	& 136	    & 11	& 74	& 12 	& 3 	        \\
Independent & 100	    & 0 	& 100 	& 0 	& 0 	        \\
\bottomrule
\end{tabular}
\vspace{-3mm}%Put here to reduce too much white space after your table 
\end{table}
\addtolength{\tabcolsep}{1.85pt}

\section{Conclusion}
We proposed a Multi-Task Federated Learning (MTFL) algorithm that builds on iterative FL algorithms by introducing private patch layers into the global model. Private layers allow users to have personalised models and significantly improves average User model Accuracy (UA). We analysed the use of BN layers as patches in MTFL, providing insight into the source of their benefit. MTFL is a general algorithm that requires a specific FL optimisation strategy, and we also proposed the FedAvg-Adam optimisation scheme that uses Adam on clients. Experiments using MNIST and CIFAR10 show that MTFL with FedAvg significantly reduces the number of rounds to reach a target average UA compared to FL, by up to $5 \times$. Further experiments show that MTFL with FedAvg-Adam reduces this number even further, by up to $3 \times$. These experiments also indicate that using private BN trainable parameters ($\gamma, \beta$) instead of statistics ($\mu, \sigma$) in model patches gives better convergence speed. Comparison to other state-of-the-art personalised FL algorithms show that MTFL is able to achieve the highest average UA given limited communication rounds. Lastly, we showed in experiments using a MEC-like testbed that the communication overhead of MTFL with FedAvg-Adam is outweighed by its significant benefits over FL with FedAvg in terms of UA and convergence speed.

\vspace{-1mm}
\bibliographystyle{ieee}
\bibliography{refs}

% \newpage 

\end{document}